\documentclass[a4paper]{jpconf}
\usepackage{graphicx}

\usepackage{hyperref}

\usepackage{amsmath}
\usepackage{amssymb}

\usepackage{algorithm}
\usepackage{algpseudocode}

\usepackage{subcaption}

\usepackage{algorithm}
\usepackage{algpseudocode}

\begin{document}

\title{Numerical optimization for Artificial Retina Algorithm \footnote{ submitted to the Journal of Physics Conference Series}}

\author{M Borisyak$^{1, 2}$, A Ustyuzhanin$^{1, 2}$, D Derkach$^{1, 2}$, M Belous$^{2}$ \\[5mm] \normalfont{on behalf of LHCb collaboration}\\[5mm]}

\address{$^{1}$ National Research University Higher School of Economics, 20 Myasnitskaya st., Moscow, Russia}
\address{$^{2}$ Yandex School of Data Analysis, 11/2, Timura Frunze St., Moscow, Russia}

\ead{mborisyak@hse.ru}

\begin{abstract}
High-energy physics experiments rely on reconstruction of the trajectories of particles produced at the interaction point. This is a challenging task, especially in the high track multiplicity environment generated by p-p collisions at the LHC energies. A typical event includes hundreds of signal examples (interesting decays) and a significant amount of noise (uninteresting examples).

This work describes a modification of the Artificial Retina algorithm for fast track finding: numerical optimization methods were adopted for fast local track search. This approach allows for considerable reduction of the total computational time per event. Test results on simplified simulated model of LHCb VELO (VErtex LOcator) detector are presented. Also this approach is well-suited for implementation of paralleled computations as GPGPU which look very attractive in the context of upcoming detector upgrades.
\end{abstract}

\section{Introduction}

Track reconstruction naturally arises in many of high-energy physics experiments: events produced by p-p collisions at the LHC energies typically include hundreds of signal examples (interesting decays) and a significant amount of noise (uninteresting examples). This makes track reconstruction a challenging task.
The substantial increase in collision energy, which leads to the increase in the number of produced tracks, makes one seek for more sophisticated event selection and reconstruction techniques which heavily rely on track finding procedures. High computational cost of event reconstruction methods gives an advantage to algorithms designed for massively parallel architectures (e.g. GPU or custom hardware). One of such algorithms is the Artificial Retina~\cite{ristori2000artificial}, a pattern-matching algorithm inspired by the structure of low-level visual recognition areas in mammal's receptive fields~\cite{hubel1962receptive}. One of the advantages of the algorithm is its extremely high parallization capacity which makes it well-suited to the track finding in high track multiplicity environments~\cite{abba2015simulation}.

In this work, we study a modification of Artificial Retina algorithm:
it is reformulated as an optimization problem and well-known methods for global optimization in continuous space are adopted. This approach allows more flexible trade-off between computational cost and track finding performance and leads to considerable reduction of total computational time.

Comparison of a grid-based and the proposed method is made on a simplified model of the LHCb VELO (VErtex LOcator) detector. The model qualitatively reflects physics in VELO, parameters for the model are inspired by parameters of Monte-Carlo simulation with the Run 3 upgrade design of the VELO~\cite{lhcb2013lhcb}.

\section{Artificial Retina algorithm}
The Artificial Retina (AR) algorithm was proposed as a fast, massively parallel track reconstruction method~\cite{ristori2000artificial}, inspired by low-level structure of line and edge detection areas of visual cortex in mammal's brain~\cite{hubel1962receptive}.
The algorithm introduces a grid of \emph{units} (or, continuing biological analogy, \emph{`cells'} or \emph{`neurons'}),
each corresponding to a particular track pattern configuration (\emph{pattern}) such as position and angle.
For each new observation (\emph{hits}) each cell computes measurements of correspondence between its pattern and the observation.

In a simple case of straight line detection in 2D space, a line can be represented by two parameters $\theta = (\alpha, \beta) \in \mathbb{R}^2$.
Thus units of AR can be arranged into a two dimensional grid, each of which corresponds to a pattern with parameters $\theta_{ij} = (\alpha_i, \beta_j)$.
Given a set of hits with coordinates $\{(x_k, y_k)\}^{N}_{k=1}$, activation (response) of a unit is typically defined as:
\begin{eqnarray}
	R(\theta) &=& \sum^{N}_{k = 1} \exp\left(-\frac{s(\mathbf{x}_k, \theta)^2}{\sigma^2}\right);\label{eq:rrf}\\
	R_{ij} &=& R(\theta_{ij});\label{eq:rr}
\end{eqnarray}
where:
\begin{itemize}
	\item $s(\mathbf{x}, \theta)$ --- the distance between hit $\mathbf{x}$ and the line defined by parameters $\theta$;
	\item $\sigma$~\textemdash~bandwidth~parameter.
\end{itemize}

As can be seen from \eqref{eq:rr}, the activation $R_{ij}$ roughly corresponds to the number of hits that are in agreement with pattern's parameters $(\alpha_i, \beta_j)$.
Typically, a set of hits aligned in a line activates a cluster of units with maximal activation in the unit with pattern parameters closest to the tracks parameters.

\begin{figure}[h]
	\centering
	\begin{subfigure}{0.45\textwidth}
		\includegraphics[width=\textwidth]{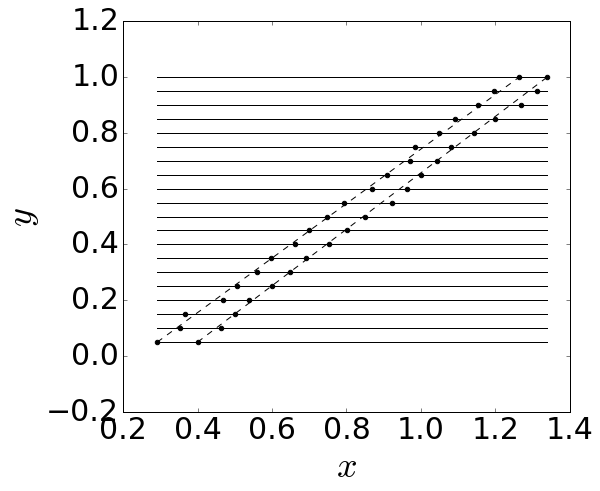}
		\caption{}
		\label{fig:closetracks}
	\end{subfigure}
	~
	\begin{subfigure}{0.45\textwidth}
		\includegraphics[width=\textwidth]{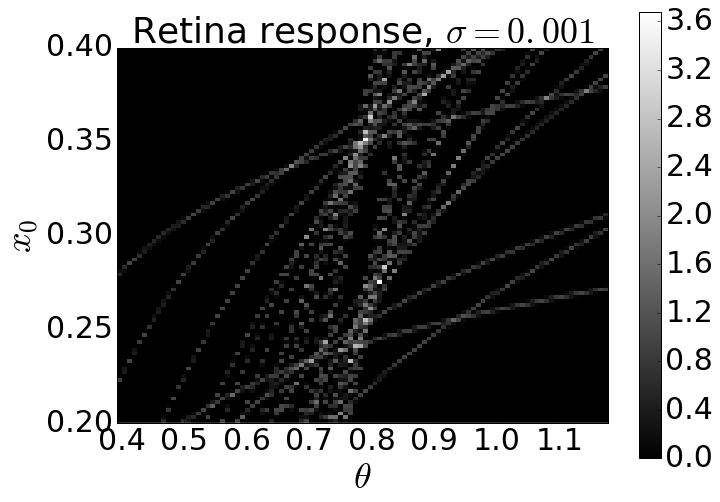}
		\caption{}
		\label{fig:smalls}
	\end{subfigure}
	\\
	\begin{subfigure}{0.45\textwidth}
		\includegraphics[width=\textwidth]{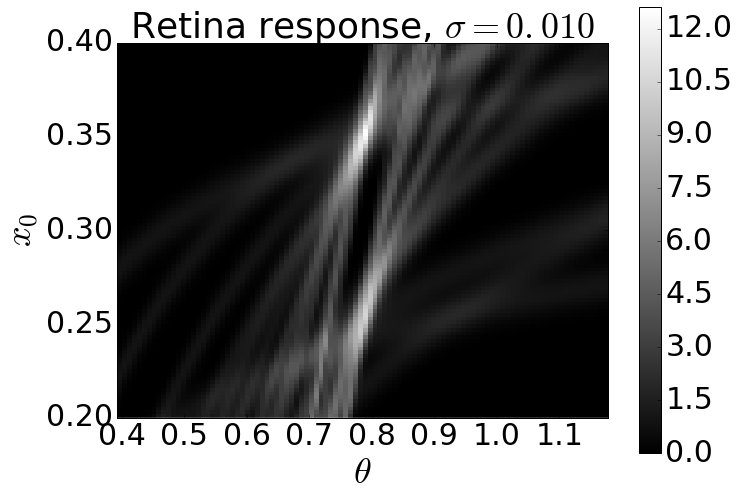}
		\caption{}
		\label{fig:norms}
	\end{subfigure}
	~
	\begin{subfigure}{0.45\textwidth}
		\includegraphics[width=\textwidth]{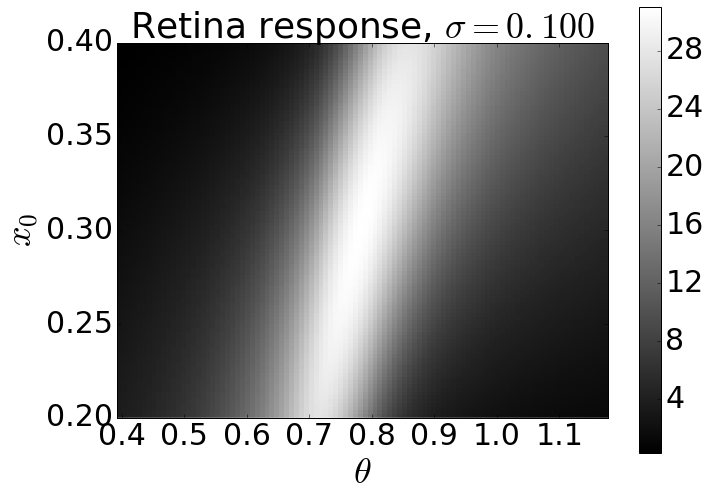}
		\caption{}
		\label{fig:bigs}
	\end{subfigure}
	\caption{
		Examples of two close tracks:
		\ref{fig:closetracks} input event with two tracks (20 hits each), noise $\epsilon_x \sim \mathcal{N}(0, 10^{-2})$
		is applied to x-coordinate of the hits; response of Artificial Retina:
		\ref{fig:smalls} for small $\sigma = 10^{-3}$;
		\ref{fig:norms} $\sigma=10^{-2}$ comparable to noise dispersion;
		\ref{fig:bigs} $\sigma=10^{-1}$, tracks are merged.
	}
	\label{fig:two_tracks}
\end{figure}

Bandwidth parameter $\sigma$ controls smoothness of the response function \eqref{eq:rr},
and should be set accordingly to the grid step and hit position errors, e.g. $\sigma \sim \Delta_\alpha = \alpha_{i + 1} - \alpha_i$.
As shown on figure \ref{fig:closetracks}, high $\sigma$ values may result in merge of two close tracks, and, when uncertainties in hits coordinates are present, low $\sigma$,
in contrast, may lead to several clusters activated by a single track.

\begin{figure}[h]
	\centering
	\begin{subfigure}{0.45\textwidth}
		\includegraphics[width=\textwidth]{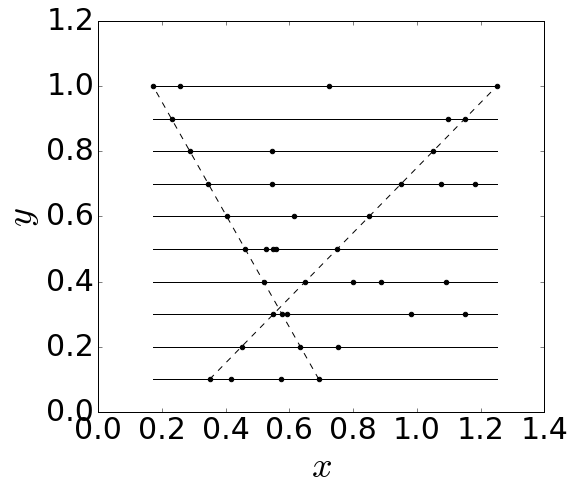}
		\caption{}
		\label{fig:data}
	\end{subfigure}
	~
	\begin{subfigure}{0.45\textwidth}
		\includegraphics[width=\textwidth]{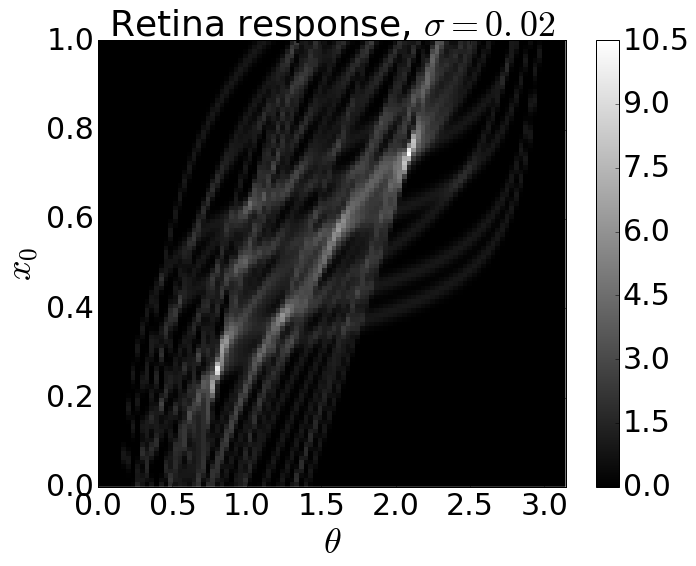}
		\caption{}
		\label{fig:rr}
	\end{subfigure}
	\caption{
		An example of Artificial Retina response: \ref{fig:data} an event with two tracks (10 hits each) and 20 uniformly distributed noise hits,
		hits are denoted by dots,
		true tracks are denoted by dashed lines,
		detector planes --- by solid lines;
		\ref{fig:rr} response of the Artificial Retina grid for $\sigma = 2 \cdot 10^{-2}$, track is parametrized by angle $\theta$ to horizontal line and offset $x_0$,
		two local maxima are close to the true track parameters, the distance between the track and the hit is defined as euclidean distance in corresponding detector planes.
	}
	\label{fig:2d}
\end{figure}

Most of the known AR algorithms rely on computing response $R_{ij}$ of the whole grid $\{\alpha_i , \beta_j\}^{N, M}_{i = 1, j = 1}$.
The usual steps of such algorithms are:
\begin{enumerate}
	\item define track model, parameter grid, distance measure; 
    \item compute responses \eqref{eq:rr} for each unit of the grid; \label{item:grid_response}
    \item select clusters of activated units;
    \item for each cluster estimate track parameters.
\end{enumerate}

One may notice that the AR algorithm can be reformulated as an optimization problem of finding all local maxima of the response function \eqref{eq:rrf} with respect to
track parameters.
From this perspective, AR algorithms described above employ brute-force approach: grid-search in parameter space.
In this work we examine one family of methods that can be used as a substitution for grid-search: first- and second-order optimization procedures.
One crucial observation is that computations of gradient and Hessian matrix of \eqref{eq:rrf} with respect to track parameters imposes relatively small overhead$^\text{\footnotemark}$,
which may bring significant benefits to the methods that can utilize this information, e.g. gradient descent.

\footnotetext{In the experiments we found that simultaneous computation of Hessian matrix, gradient and response functions takes only around 3 times more time than response function alone.}

However, the problem of finding all local maxima of the response function \eqref{eq:rrf} is intrinsically non-convex and non-local, hence global optimization strategies must be adopted.
One of such strategies is the multi-start algorithm~\cite{marti2003multi}, which
allocates $q$ initial guesses drawn from prior distribution $P_\theta$ and then sequentially updates each of them.
In this study the number of updates for each initial guess is fixed.

Pseudo-code for the proposed method is shown in listing~\ref{alg:fast_retina}: function $\mathrm{update}$ denotes selected optimization procedure.

\begin{algorithm}[h]
	\caption{Accelerated Artificial Retina algorithm}
	\label{alg:fast_retina}
	\begin{algorithmic}
		\Function{accelerated-artificial-retina}{$\mathbf{X}$, $n$, $q$}
			\For{$i = 1, \dots, n$}
				\State draw $\theta^0_i$ from prior distribution $P_\theta$
			\EndFor
			\For{$j = 1, \dots, q$}
				\For{$i = 1, \dots, n$}
					\State compute response $R(X, \theta^{j - 1}_i)$, gradient $\nabla R(X, \theta^{j - 1}_i)$ and Hessian $\mathbf{H} R(X, \theta^{j - 1}_i)$
					\State $\theta^j_i := \mathrm{update}\left[R(X, \theta^{j - 1}_i), \nabla R(X, \theta^{j - 1}_i), \mathbf{H} R(X, \theta^{j - 1}_i)\right]$ \label{line:update}
				\EndFor
			\EndFor
			\State cluster solutions $\{ \theta^q_i \}^n_{i = 1}$
			\State for each cluster select solution with highest response and above threshold $R_0$
			\State \Return selected solutions
		\EndFunction
	\end{algorithmic}
\end{algorithm}

\section{Simplified LHCb VELO model and experiment}
\label{sec:experiment}
We illustrate the application of Artificial Retina for tracking on the example of a simplified model of the LHCb Vertex Locator (VELO) detector.
This simplified model (sVELO) is inspired by the VELO upgrade Technical Design Report~\cite{lhcb2013lhcb}, and is aimed to capture all VELO details essential from the tracking point of view.

At LHCb two crossing beams result in proton-proton collisions which produce numerous secondary particles.
The collision  point (called primary vertex) is far from the magnet, thus secondary particles trajectories can be considered as straight lines.
The VELO detector surrounds interaction region and consists of $N_l$ layers perpendicular to the beam axis ($z$-axis).
Each layer consist of a number of silicon detectors (pixels) that react on a charged particles crossing the material.

An event consist of $N_e$ coordinates $(x, y, z)$ of triggered pixels (hits): either activated by a secondary particle, or noise.

In sVELO we assume that:
\begin{itemize}
	\item number of layers $N_l = 20$;
	\item each layer is a disk with radiuses: outer $r_{\mathrm{outer}} = 42 \text{mm}$, inner $r_{\mathrm{inner}} = 8 \text{mm}$;
	\item layers are equally spaced within 700 mm along $z$-axis;
	\item particles are travelling in straight lines;
	\item pseudo-rapidity of particles $\eta = -\ln \left[\tan\left(\frac{\theta}{2}\right)\right]$ is distributed uniformly $\eta \sim U[1, 6]$;
	\item angle in the traverse plain $\phi$ is distributed uniformly: $\phi \sim U[0, 2 \pi]$;
	\item each particle has a probability $p_{\mathrm{hit}} = 0.5$ of interacting with detector layer;
	\item particles that leave less than $N_{\min} = 2$ hits are considered as undetectable and their hits are marked as noise;
	\item errors of $(x, y)$ coordinate measurements are distributed normally: $\epsilon_x, \epsilon_y \sim \mathcal{N}(0, 10^{-2})$;
	\item $N_\mathrm{noise}$ uniformly distributed hits are introduced in each event with $N_\mathrm{noise} \sim \mathrm{Poisson}(250)$.
\end{itemize}

In the experiment number of reconstructible tracks was varied between 50 and 350.

The track is parametrized by two angles $\theta$ and $\phi$:
\begin{eqnarray*}
	x(t) &=& t \sin \theta;\\
	y(t) &=& t \cos \theta \sin \phi;\\
	z(t) &=& t \cos \theta \cos \phi.
\end{eqnarray*}
The distance function $s$ is defined as euclidean distance between the hit and intersection of a track within hit's detector layer.

A track is considered to be reconstructed if the algorithm reports an estimation within $\epsilon = 10^{-3}$ radians from true track's parameters
(which is comparable to the angular size of VELO pixel).

In the experiment we found that for this distance function routine for computing AR response, its gradient and Hessian matrix takes
less than 3 time longer than computations of response alone ($T_0$), normalization constant $C = 3$ is used to account the time difference for
different routines. To show the reduction in total computational time relative to plain grid-search, we set the number of initial guesses, so that computational resources are $\alpha$ fraction of these required by grid-search:
\begin{equation}
	n = \alpha \cdot n_{\mathrm{grid}} \frac{1}{C_0}\frac{1}{q}
\end{equation}
where:
\begin{itemize}
	\item $n_{\mathrm{grid}}$ --- number of cells required for plain grid-search to provide $\epsilon$ resolution,
	\item $q = 3$ number of optimization steps,
	\item $C_0$ --- time required by each optimization step normalized by $T_0$.
\end{itemize}

Among all methods examined during the experiment, the Truncated Newton method\cite{nocedal2006numerical, nash1984newton} was found to yield the best results.
For this methods the normalization constant $C_0 \approx 30$.
During the experiments we discovered that slight improvement of the results can be achieved by updating bandwidth parameter $\sigma$ with each optimization step,
sequence $\sigma_1~=~0.3, \sigma_2~=~0.175, \sigma_3~=~0.05$ was used in this study.

\begin{figure}[h]
	\centering
	\begin{subfigure}{0.45\textwidth}
		\includegraphics[width=\textwidth]{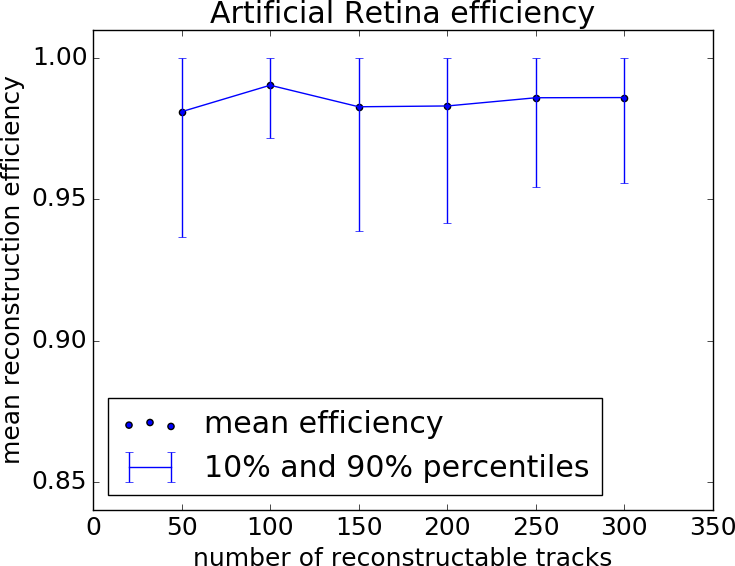}
		\caption{Efficiency for $\alpha = \frac{1}{3}$}
		\label{fig:third}
	\end{subfigure}
	~
	\begin{subfigure}{0.45\textwidth}
		\includegraphics[width=\textwidth]{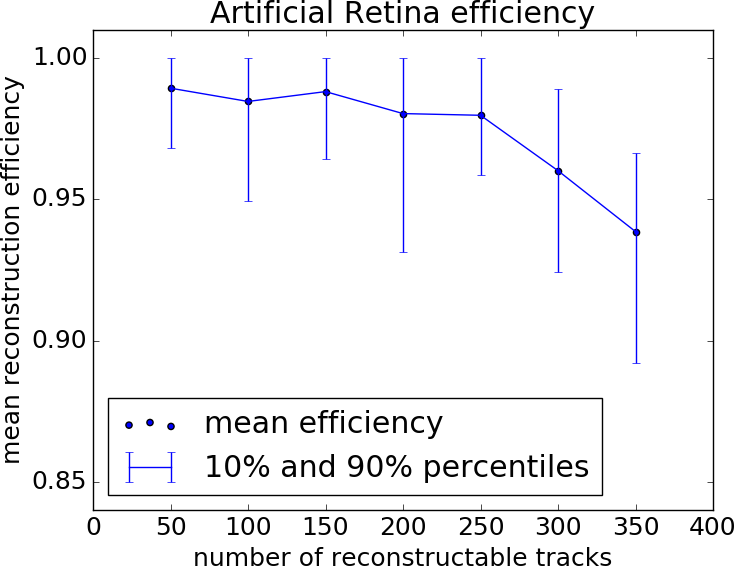}
		\caption{Efficiency for $\alpha = \frac{1}{10}$}
		\label{fig:tenth}
	\end{subfigure}
	\caption{
		Experimental results for $\alpha = \frac{1}{3}$ and $\alpha = \frac{1}{10}$.
		Horizontal axis correspond to number of reconstructible tracks, vertical --- efficiency,
		fraction of reconstructed tracks.
	}
	\label{fig:results}
\end{figure}

Results are shown in figure \ref{fig:results}. Generally, the efficiency of the algorithm is high.
From the figure \ref{fig:tenth} it can be clearly seen that the efficiency of the algorithm decreases as number of initial seeds approaches
to the number of tracks. Nevertheless, with enough initial seeds (figure \ref{fig:third}) efficiency is close to 1, while the whole procedure requires only one third of the resources required by grid-search method.

\section{Conclusion}

In this work we examined a modification of the Artificial Retina algorithm that adopts continuous space optimization methods and
multi-start procedure. High convergence rates of these methods overcome gradient and Hessian matrix computational costs, which results in
overall reduction of total computation time.

Experiments on a simplified model of LHCb VErtex LOcator detector were performed and showed that hat
it is possible to keep track reconstruction efficiency above 95\% and thanks to the method proposed the computational time can be reduced by the factor of 3 compared to the grid-search based
Artificial Retina algorithm.

\section*{References}
\bibliographystyle{iopart-num}
\bibliography{iopart-num.bib}

\providecommand{\newblock}{}
\begin{thebibliography}{1}
\expandafter\ifx\csname url\endcsname\relax
  \def\url#1{{\tt #1}}\fi
\expandafter\ifx\csname urlprefix\endcsname\relax\def\urlprefix{URL }\fi
\providecommand{\eprint}[2][]{\url{#2}}

\bibitem{ristori2000artificial}
Ristori L 2000 {\em Nuclear Instruments and Methods in Physics Research Section
  A: Accelerators, Spectrometers, Detectors and Associated Equipment\/} {\bf
  453} 425--429

\bibitem{hubel1962receptive}
Hubel D~H and Wiesel T~N 1962 {\em The Journal of physiology\/} {\bf 160}
  106--154

\bibitem{abba2015simulation}
Abba A, Bedeschi F, Citterio M, Caponio F, Cusimano A, Geraci A, Marino P,
  Morello M, Neri N, Punzi G {\em et~al.\/} 2015 {\em Journal of
  Instrumentation\/} {\bf 10} C03008

\bibitem{lhcb2013lhcb}
{LHCb Collaboration and others} 2013 {LHCb VELO upgrade technical design
  report} Tech. rep.

\bibitem{marti2003multi}
Mart{\'\i} R 2003 Multi-start methods {\em Handbook of metaheuristics\/}
  (Springer) pp 355--368

\bibitem{nocedal2006numerical}
Nocedal J and Wright S~J 2006 Numerical optimization 2nd

\bibitem{nash1984newton}
Nash S~G 1984 {\em SIAM Journal on Numerical Analysis\/} {\bf 21} 770--788

\end{thebibliography}

\end{document}